\documentclass[letterpaper, 10 pt, conference]{ieeeconf}

\IEEEoverridecommandlockouts                              
\usepackage{times}

\usepackage{graphicx}

\usepackage[sorting=none]{biblatex}
\AtBeginBibliography{\small}
\usepackage{multicol}
\usepackage{threeparttable}
\usepackage{colortbl}
\usepackage{multirow}
\usepackage[bookmarks=true]{hyperref}
\usepackage{hyperref}
\usepackage{algorithm,algpseudocode}
\usepackage{amsmath}
\usepackage{balance} 
\usepackage{capt-of,etoolbox}

\usepackage{graphics} 
\usepackage{amssymb}  
\usepackage{colortbl}
\usepackage{color}
\usepackage{mathtools}
\usepackage{subfigure}
\usepackage{amsmath,amssymb,amsfonts}
\usepackage{amssymb}  
\usepackage{graphicx}
\usepackage[dvipsnames]{xcolor}
\usepackage{tabularx}
\usepackage{multirow}
\usepackage{colortbl}
\usepackage{color}
\usepackage{flushend}

\newcommand{\loosepar}{\looseness=-1}

\title{\LARGE \bf Human Stress Response and Perceived Safety during  Encounters with Quadruped Robots
}

\author{Ryan Gupta$^{*,1}$, Hyonyoung Shin$^{*,1,2}$, Emily Norman$^{3}$, Keri K. Stephens$^{3}$, Nanshu Lu$^{1,2}$ and Luis Sentis$^{1}$
\thanks{$^{*}$These authors contibuted equally to this work.}
\thanks{$^{1}$Department of Aerospace Engineering and Engineering Mechanics,
        University of Texas at Austin,
        Austin, TX 78712 USA
        {\tt\small ryan.gupta@utexas.edu}}%
\thanks{$^{2}$Department of Electrical and Computer Engineering, 
               University of Texas at Austin, 
               Austin, TX 78712 USA}
\thanks{$^{3}$Department of Communication Studies,
        University of Texas at Austin,
        Austin, TX 78712 USA
}}

\begin{document}


\maketitle
\thispagestyle{empty}
\pagestyle{empty}

\begin{abstract}
Despite the rise of mobile robot deployments in home and work settings, perceived safety of users and bystanders is understudied in the human-robot interaction (HRI) literature. To address this, we present a study designed to identify elements of a human-robot encounter that correlate with observed stress response. Stress is a key component of perceived safety and is strongly associated with human physiological response. In this study a Boston Dynamics Spot and a Unitree Go1 navigate autonomously through a shared environment occupied by human participants wearing multimodal physiological sensors to track their electrocardiography (ECG) and electrodermal activity (EDA). The encounters are varied through several trials and participants self-rate their stress levels after each encounter. The study resulted in a multidimensional dataset archiving various objective and subjective aspects of a human-robot encounter, containing insights for understanding perceived safety in such encounters. To this end, acute stress responses were decoded from the human participants' ECG and EDA and compared across different human-robot encounter conditions. Statistical analysis of data indicate that on average (1) participants feel more stress during encounters compared to baselines, (2) participants feel more stress encountering multiple robots compared to a single robot and (3) participants stress increases during navigation behavior compared with search behavior.\loosepar{}


\end{abstract}


\section{Introduction}
Autonomous mobile robot research has been ongoing for decades for a wide range of real-world scenarios like delivery \cite{lee2021assistive}, search \cite{gupta2023learned}, surveillance \cite{grocholsky2006cooperative} and mapping \cite{charrow2015information}. 
As mobile robots proliferate in communities, designers must consider the impacts these systems have on the users, onlookers and places they encounter. 
It becomes increasingly necessary to study situations where humans and robots coexist in common spaces, even if they are not directly interacting.\loosepar

Subjective questionnaires have been popularly employed in literature to measure preceived safety, preference and human mental state during HRI \cite{bartneck2009measurement,hauser2023s,van2023increasing,taylor2022observer}.  
However, questionnaires have poor temporal resolution and are subject to many sources of bias including moderacy response bias and memory effects induced by repetitive administration \cite{choi2005peer}. 
Questionnaires also impact HRI, requiring researchers to adopt rapidly administered questionnaires which have limited detail and coverage or, in some cases, limit the choice of robots to be studied to those like the Pepper robot with a user interface for questionnaires \cite{akalin2023taxonomy,rubagotti2022perceived}. 
To overcome these limitations, physiological measures have also been considered in HRI \cite{dehais2011physiological,chen2011touched,stark2018personal,willemse2017affective}. 
Non-invasive physiological sensing enables the tracking of the human internal state in real-time, for example to evaluate stress during HRI \cite{akalin2022you}. \loosepar 

\begin{figure}
    \centering
    \vspace{-1em}
    \includegraphics[width=\linewidth]{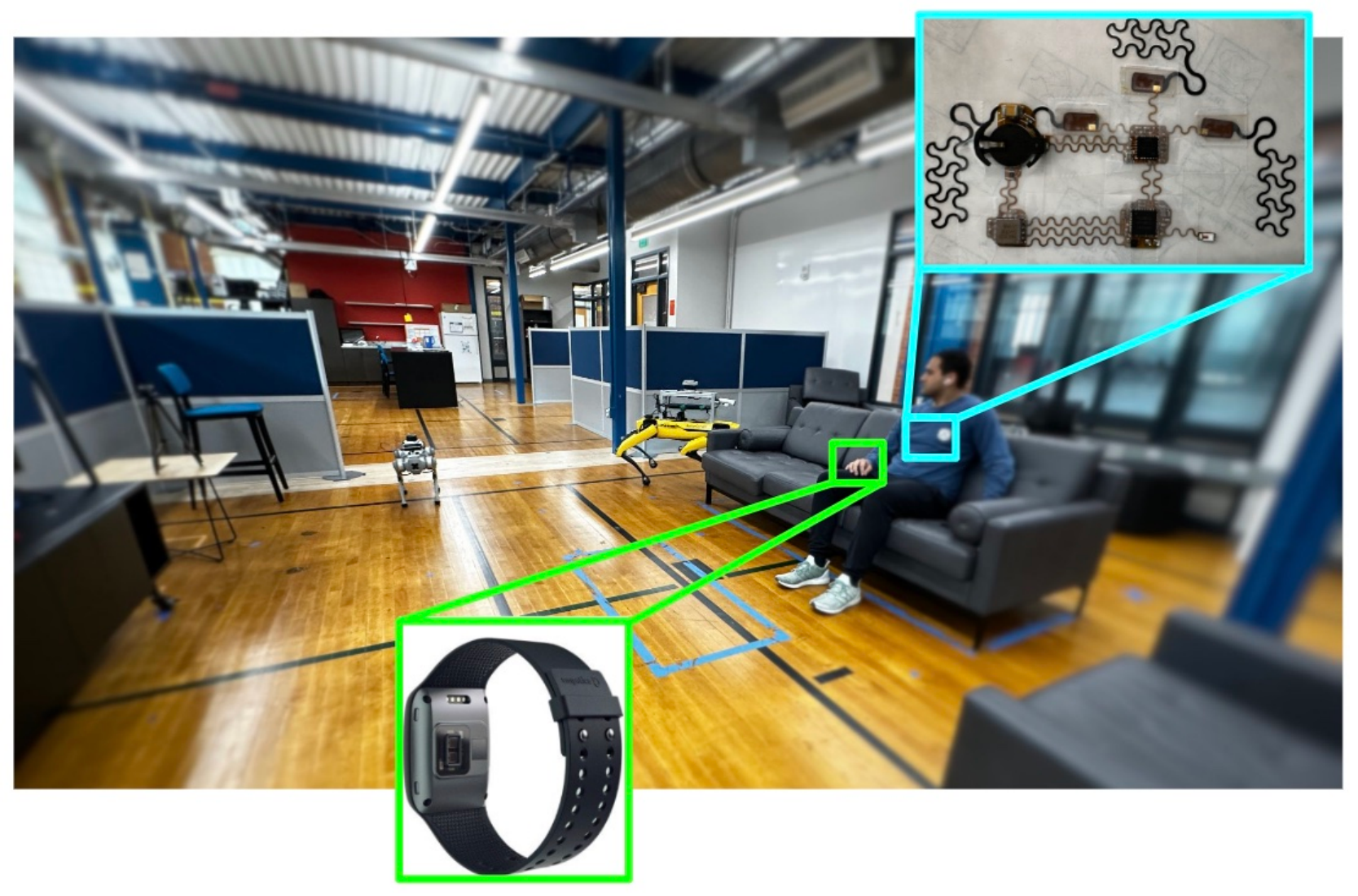}
    \vspace{-1em}
    \caption{An overview of the experiments. The Unitree Go1 and Boston Dynamics Spot perform various mobile behaviors in shared spaces with UT community members. Physiological measures are collected from the E4 wristwatch (green) \cite{e4} and chest e-tattoo (cyan) \cite{bhattacharya2023chest}.}
    \label{fig:gcr-header}
    \vspace{-2em}
\end{figure}
We hypothesize these sensors allow tracking perceived safety during human-robot encounters by decoding acute stress levels from known biomarkers in ECG and EDA data. 
This is significant in two ways: (1) To the best of our knowledge, this is the first analysis of physiological response to mobile robots navigating freely \cite{rubagotti2022perceived}. 
Our study deploys two quadruped robots in a realistic apartment setting (Fig. \ref{fig:gcr-header}). 
(2) We characterize \textit{human-robot encounters}.
This subset of human-robot interactions involve humans and robots coexisting in a shared space with awareness of one another, who are not actively interacting.
Human-robot encounters represent situations that are likely to dominate total interaction time if mobile robots are deployed in home or other indoor environments. 
Our results provide a few insights, based on physiological signal analysis, regarding the elements of human-robot encounters that affect the perceived safety of the robots by humans. 
These insights can be used in future design where perceived safety is a concern and pave the way for an online pipeline whereby the signals from a user's sensors could be decoded to inform robot behavior in real-time \cite{staffa2022enhancing,rani2004anxiety}.\loosepar

\section{Background}

\subsection{Stress Response and Comfort in HRI}


Perceived safety during encounters with mobile robots remains understudied in many domains \cite{hauser2023s,akalin2023taxonomy}.
The first group of works we find leverage more objective physiological measures to identify stress response.
These studies integrate robots in human activities with known external stressors like seminar style presentations or cognitive quizzes to evoke discomfort \cite{akalin2022you,staffa2022enhancing,turner2019use}.
They ask how replacing humans with robots impacts people's response to situations, however none consider mobile robots moving freely through shared environments.
The second group of studies leverage more subjective measures, for example statistical analysis of subjective questionnaires \cite{hauser2023s,van2023increasing,aliasghari2023we,wang2023effects,cohen1994perceived,mavrogiannis2022social} or qualitative evaluation of open form responses \cite{doring2019love,fortunati2021rise,joffe2011thematic,terry2017thematic}.
Ultimately, the proposed study differentiates from many works, using a combination of methods to investigate perceived safety during mobile robot encounters.
\loosepar

\begin{figure}
    \centering
    \includegraphics[width=\linewidth]{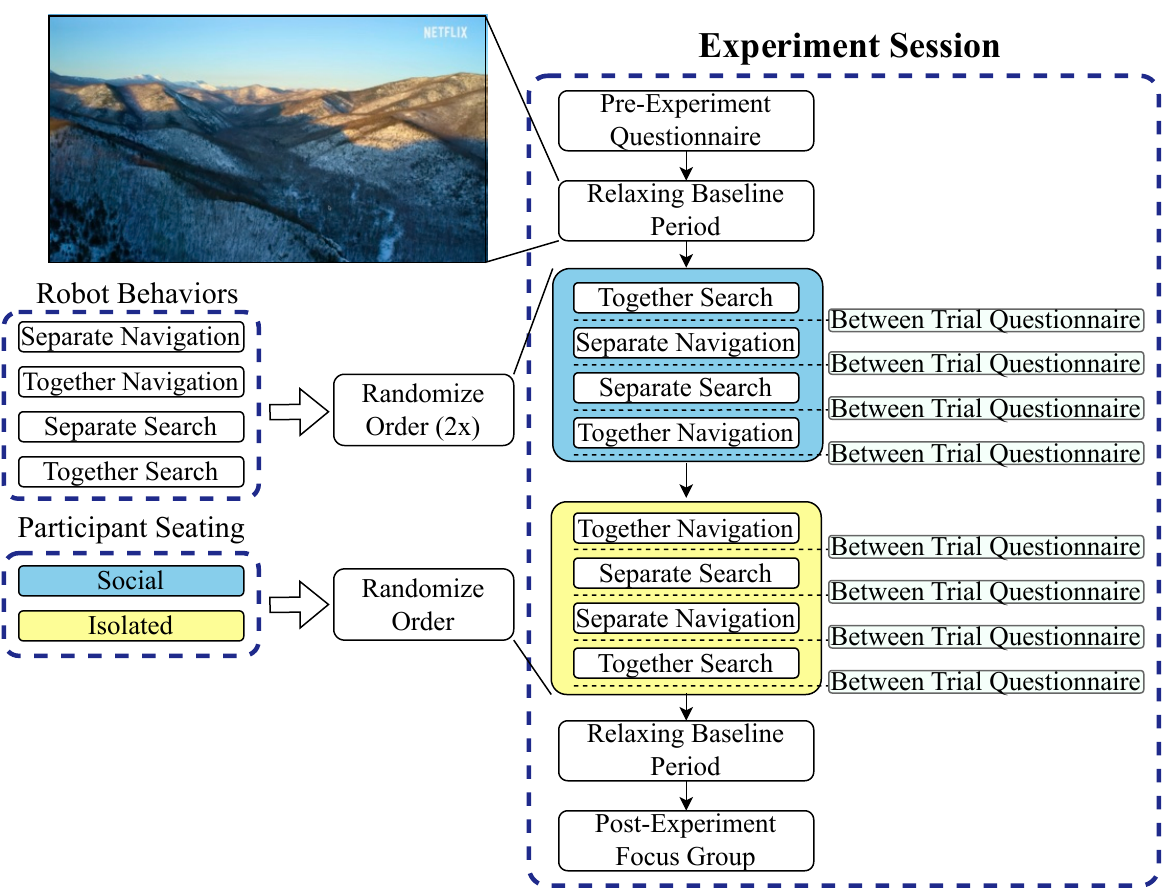} 
    \vspace{-2em}
    \caption{Overview of an example experiment session designed to gain insights into human responses to encounters with robots. On the left are the list of 4 robot behaviors and 2 participant seating options. The order of each are randomized and then inserted into the experiment session as shown on the right side. The first step is receiving consent from the participants. The nature video is repeated before and after the experimental session as a relaxed baseline phsyiological state. After sensors are removed, participants take part in an interview.\loosepar}
    \vspace{-2em}
    \label{fig:experiment-overview}
\end{figure}

\subsection {Stress Decoding from Physiological Signals} \label{ssec:physiosignals}
Non-invasive physiological recordings have been shown to contain a wealth of information about human physical and mental states \cite{greco2021acute,thehrvreview}. 
We are interested in mental stress, particularly acute stress from short recordings. 
This is fulfilled by electrodermal activity (EDA) recordings, or galvanic skin response, which measures changes in skin conductance over time on the palm, foot, or wrist \cite{zangroniz2017electrodermal,critchley2002electrodermal}.
Skin conductance time series consists of two components: tonic and phasic. Tonic refers to the low-frequency changes in the baseline of the signal, while the phasic refers to the sharp spikes followed by recovery to baseline in the signal \cite{greco2015cvxeda}. 
The phasic component is known to correspond to psychological arousal events, allowing quantitative determination of mental stress over time. 
Decoding acute stress from tonic-phasic decompositions of EDA signals has been demonstrated extensively \cite{greco2021acute,zangroniz2017electrodermal,critchley2002electrodermal,caruelle2019use}.\loosepar

Another signal to detect mental stress is heart rate variability (HRV) \cite{kim2018stress}, which can be evaluated from any modality for reliable heart beat information. 
HRV features have been linked to a variety of both acute and chronic stress in both real life and experimentally induced situations \cite{thehrvreview}. 
This is due to the fact that HRV reflects complex changes in sympathetic and parasympathetic nervous systems which manage the homeostasis of psychological arousal. 
HRV is often evaluated over long recordings; however it can be reliably estimated from ultra-short-term and short-term recordings like our study \cite{USTHRV1,USTHRV2}.
Acute stress has been associated with decreases in HRV entropy in both experimentally induced situations \cite{USTHRV2,brugnera_entropy} and real-life situations \cite{dmitri_entropy,exam_entropy}. 
SDNN (Standard Deviation of NN intervals) and RMSSD (Root Mean Square of Successive Differences) are both measures commonly used in the analysis of heart rate variability (HRV), which is the variation in the time intervals between successive heartbeats.
Decreases in SDNN and RMSSD and increases in low-frequency bandpower (0.04–0.15 Hz) compared to high-frequency (0.15–0.40 Hz) are also commonly noted features of acute stress \cite{thehrvreview}.  \loosepar

Estimation of mental stress from EDA and HRV is non-trivial due to these signals being indirect measures of certain changes caused by the autonomic nervous system, acting as proxies for mental stress.
Therefore, to maximize success, raw signals from participants must be as high-quality and free of noise and motion artifacts as possible, to ensure that no information is lost to such factors. 
While this incentivizes the use of clinical grade equipment, this conflicts with our study goals where we are interested in realistic human-robot encounters in an apartment setting. 
We conclude that state-of-the-art wearable devices are more suitable to avoid lengthy sensor set-up, skin preparation and cables that interfere with movement or cause stress on their own.
To this end, for ECG (HRV), we used a wireless chest e-tattoo device \cite{bhattacharya2023chest} which is accurate while being minimally intrusive due to its ultrathin and lightweight form factor. Although this device records both electrocerdiography (ECG) and seismocardiography (SCG), only the former was used for analysis in this study. For EDA, we used an Empatica E4 wristwatch \cite{e4}, which, according to a 2023 review on wearables used to assess perceived stress \cite{E4popularity}, was used in 56.3\% of such studies. \loosepar

\begin{figure*}[h!]
    \centering
    \includegraphics[width=0.975\linewidth]
    {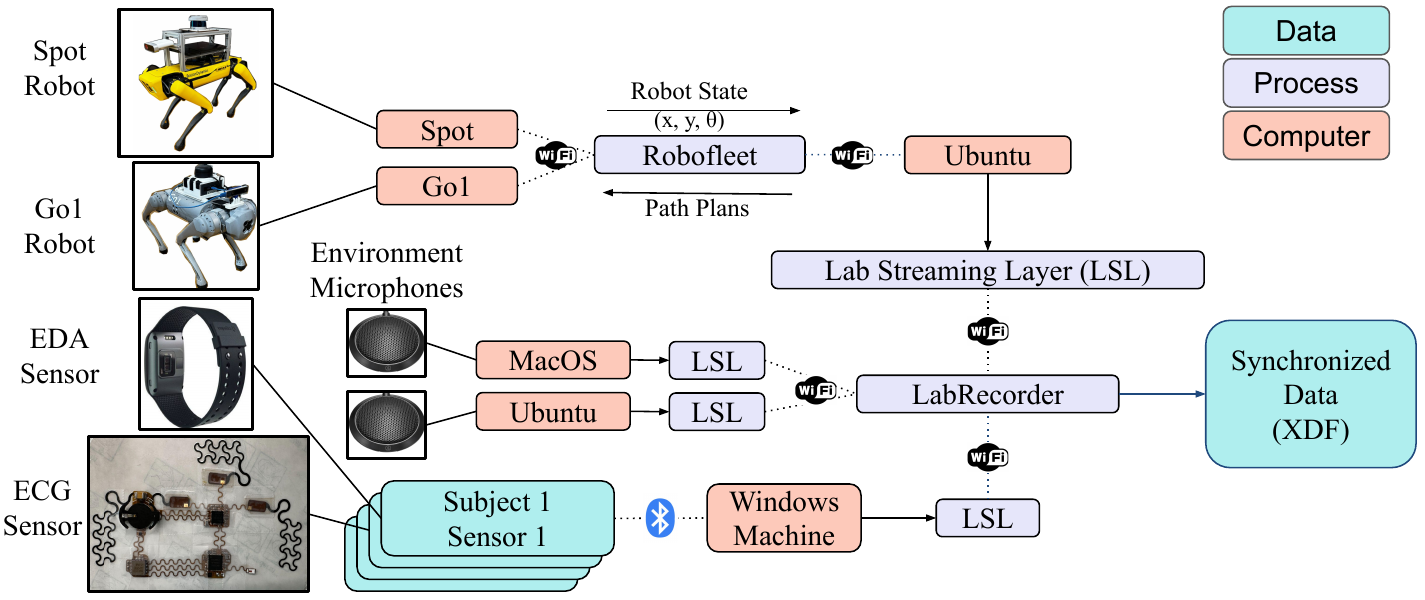}
    \vspace{-1em}
    \caption{A figure demonstrating the syncronized data acquisition architecture. The Spot and Go1 robots send state information via Robofleet \cite{sikand2021robofleet} to an Intel NUC base station. Two omnidirectional microphones connect to a MacBook Pro (MBP) and Ubuntu (knapsack) machines. Finally, the two sensors that was participants wears connect via Bluetooth to designated Android and Windows machines. LabStreamingLayer \cite{lsl} is used to push timestamped sensor data from all different sources to the network. LabRecorder \cite{labrecorder} is used to record all streams into a single synchronized XDF file.\loosepar }
    \vspace{-1em}
    \label{fig:sync-daq}
\end{figure*}

\section{Experiment Design}

We conduct a study with 17 total participants over 9 sessions to investigate perceived safety in human robot encounters. 
This study was independently reviewed and approved by the University of Texas Institutional Review Board (IRB \#00004099).
An overview of experimental sessions are shown in Fig. \ref{fig:experiment-overview}.
The blockwise randomization of the trial configurations ensured that the effect of confounding factors such as time were removed. Fig. \ref{fig:experiment-overview} shows each configuration was also repeated twice to enhance the repeatability of results from each configuration.
The research questions in this work are as follows:\loosepar
\begin{itemize}
    \item $Q1:$ Do participants have a stress response when robots are present compared to baseline sessions?
    \item $Q2:$ What mobile robot dog behaviors cause human stress response?
    \item $Q3:$ Do participants feel more comfortable when seated together vs. when isolated?
\end{itemize}


\subsection{Data Acquisition}

Synchronized data acquisition from all physiological sensors and the robots enables analysis of response to the varied conditions.
We also collect data from two microphones in the environment, one in the small room and one in the large living room near the participant seating locations that record audio intensity streams.
From each robot we collect position and orientation, represented as $(x, y, \theta)^{Spot}_t$ and $(x, y, \theta)^{Go1}_t$ recorded at time $t$.
Each participant also wears the chest e-tattoo and the Empatica E4 wristwatch.\loosepar

Figure \ref{fig:sync-daq} shows the architecture for real-time synchronized data acquisition.
Robot data is sent as timestamped ROS \cite{ros.org} messages via Robofleet \cite{sikand2021robofleet} to a base station.
The base station and other systems stream robot and microphone data via LabStreamingLayer \cite{lsl}.
EDA and ECG sensors on each participant connect via Bluetooth to a Windows machine for streaming.
All data is recorded using LabRecorder \cite{labrecorder} and saved as a synchronized Extensible Data Format (XDF) file.\loosepar

\subsection{Conditions}

\begin{figure}[]
    \centering
    \includegraphics[width=\linewidth]{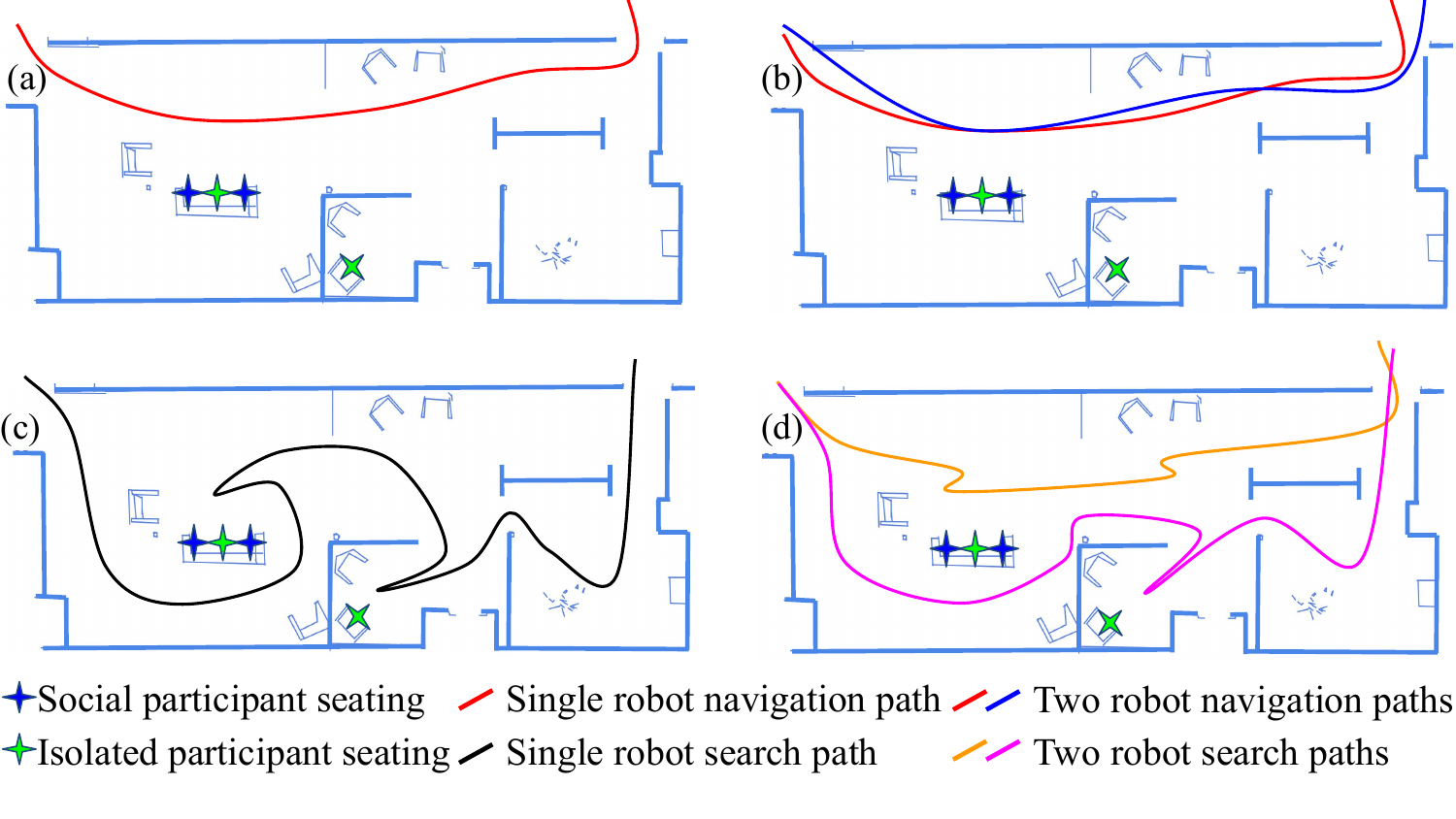}
    \vspace{-2em}
    \caption{An overview of the three control variables. Namely, four robot behaviors and two participant seating settings. In (a) the social and isolated participant seating locations are shown with the single robot navigation path. (b) is the two robot navigation paths. In (c) is the single robot search path and (d) shows the two robot search paths.}
    \vspace{-1em}
    \label{fig:conditions}
\end{figure}

We control three variables: (1) single v. multi-robot, (2) navigation v. search behavior, and (3) social v. isolated participant seating.
Fig. \ref{fig:conditions} demonstrates the control variables overlaid on a map of the apartment, showing the four robot behaviors.
The figure shows the navigation paths Fig. \ref{fig:conditions}(a,b) and the search paths Fig. \ref{fig:conditions}(c,d). 
In the large room, one or both participants sit together on a couch and then one or no participant on a chair in the small room (Fig. \ref{fig:conditions}(a)).
A pre-experiment questionnaire collects data about the personality, baseline stress and demographics of the involved participants.\loosepar 


\subsection{Experiment Procedure}
First, participants are presented with a consent form. 
Upon signing consent, the chest e-tattoo and E4 wristwatch are applied to participants.
Participants are seated together on the living room couch to fill out a pre-experiment questionnaire. 
The participants then watch a 5 minute relaxing nature video to establish a relaxed baseline state for physiological data. 
After, the participants' seating locations are updated to isolated if necessary (chosen at random).
Next, experimental trials begin, each characterized by (1) a seating position, (2) single or multi-robot and (3) search or navigation behavior.
Order effects of the conditions are addressed by counterbalancing.
In particular, we randomize the isolated v. participant seating and each of the four robot behaviors within each seating condition. 
Between trials participants take a 10 question Likert scale questionnaire. 
It is a combination of surveys focused on stress, comfort and perceived safety used in \cite{akalin2022you}.
After four trials, participant seating is switched.
The same 4 behaviors are again repeated in the new seating locations with order again randomized.
Participants then watch the same 5 minute nature video, establishing a second known baseline relaxation state to confirm that perturbations in the physiological signals during the 8 trials were induced by the robot and not a confounding factor such as elapsed time in the experiment and fatigue.
After the nature video, participants removed the sensors and were debriefed then interviewed.\loosepar

\section{Data Analysis Methods}
\label{sec:data-analysis}

\subsection{Signal Processing and Feature Extraction}
ECG signals were validated first by visually examining the waveform of the average R peak grand average for a clear QRS complex and then by ensuring that the heart rate was reasonable. 
HRV features were extracted from peak information using Neurokit2 \cite{makowski2021neurokit2}, which returns a total of 91 features from both the time and frequency domains as well as non-linear features.
From these, 26 features cannot be reliably evaluated from ultra-short-term recordings and were removed. 
In our case, this consisted of HRV bandpower in the ULF and VLF bands, detrended fluctuation analysis (DFA)-related indices, as well as any time domain features that required recordings longer than 5 minutes. 

We found that there was a level of noise mixed into the EDA signals in some sessions, possibly due to poor skin-electrode contact during subject motion and the low sampling rate of the device. 
To mitigate this, a 2nd order Savitzky–Golay filter was used to act as a low-pass filter and smoothen out the noise before passing the pre-processed signal to the cvxEDA algorithm \cite{greco2015cvxeda}. 
This algorithm decomposes the 1-dimensional EDA signal into phasic and tonic components via convex optimization. EDA features were then calculated according to the methodology in \cite{greco2015cvxeda}, with the addition of EDA autocorrelation \cite{autocor}, normalized EDA sympathetic index, as well as median and maximum amplitudes of the phasic sudomotor nerve activity (SMNA) peaks for a total of 15 EDA features. Features were normalized by the recording length wherever necessary. 
EDA and HRV features were then Z-scored and combined to result in a total feature matrix of 17 subjects $\times$ 10 trials $\times$ 80 features. \loosepar

\subsection{Exploratory Data Analysis} \label{ssec:explore}
Next, we used the Chi-squared test to examine the 20 most statically significant features for the Baseline v. Robot condition. 
We chose a Chi-squared test compared with a principal component analysis (PCA) for interpretability of the results.
While a PCA can be performed for a similar purpose, the resulting components cannot be compared with features of acute stress identified from literature as done in this work, which results in less certainty about the applicability of the results.
This is due to the fact that there is no common ``stress metric" in literature which can be used. 
In the feature selection via chi-squared test function available in Matlab, which is good to use for continuous data for classification problems, the features are ranked according to negative logs of the p-values.
We note that EDA and HRV features were shortlisted in an approximately even proportion (9 EDA, 11 HRV) by the algorithm despite more candidate HRV features evaluated. This suggests that the multimodal approach based on both cardiac and electrodermal activity was informative for the estimation of robot-induced stress. Since exposure to robots is not a well-established stressor condition (in contrast to established psychological or neuroscientific tasks such as Stroop and Speech tests), an important step is to: (1) further examine the shortlisted features based on whether they have reasonable physiological background for being predictive of acute stress and arousal, and (2) examine the evolution of these features in the encounter conditions of interest. These steps will help to ensure that the predictive performance of the automatic human-robot encounter classifier in Section \ref{sssec:ml} is not a result of overfitting. \loosepar

\begin{figure}
    \centering
    \includegraphics[width=\linewidth]{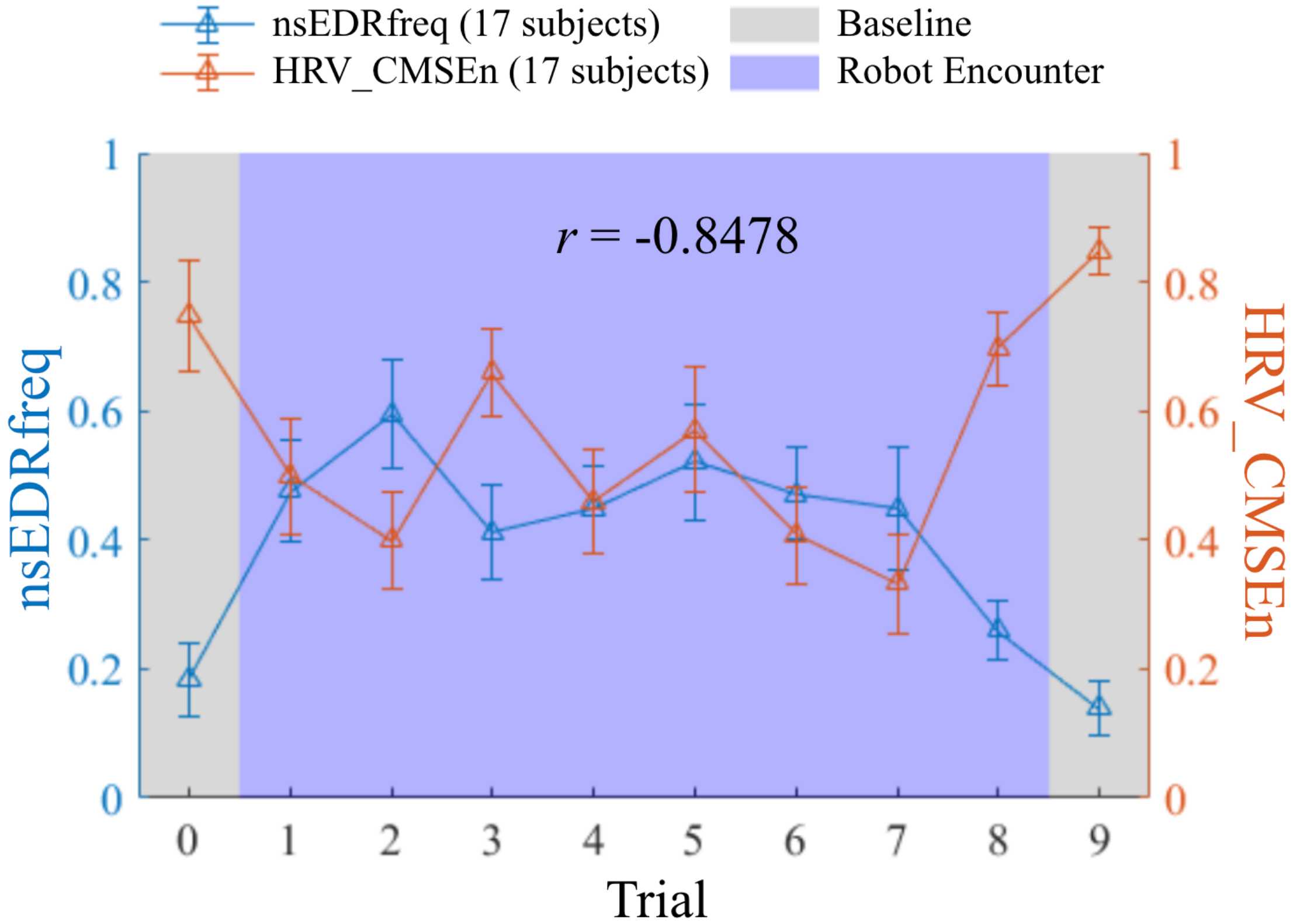}
    \vspace{-2em}
    \caption{Normalized features HRV\textsubscript{CMSEn} and nsEDRfreq over trials i.e. experiment time. Data shows averaged data from 17 subjects and their standard error. The trends show negative correlation as expected ($r = -0.8478$, see also Section \ref{ssec:explore}) and show a clear U-shape as values representing the baseline relaxed state are perturbed by the introduction of robots into the environment in trials 1 through 8 before returning to normal (see further analysis in Fig. \ref{fig:principal_features}).\loosepar }
    \vspace{-2em}
    \label{fig:over_trial}
\end{figure}

\begin{figure*}[t]
    \centering
    \includegraphics[width=\textwidth]{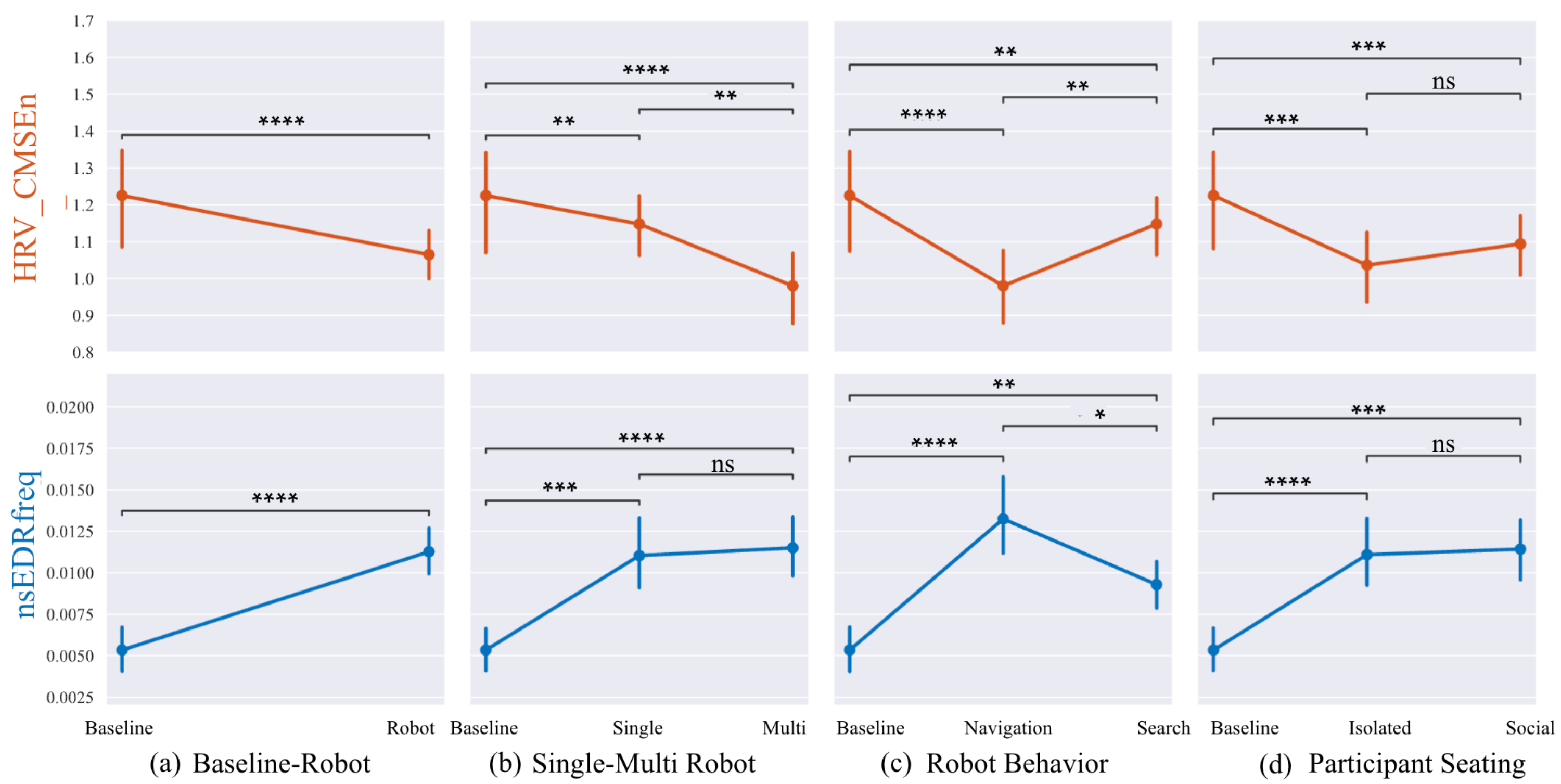}
    \vspace{-2em}
    \caption{Statistical comparison for the principal features HRV\textsubscript{CMSEn} and nsEDRfreq in each possible experimental condition, with all subject data. Error bars show the 95\% confidence interval. Asterisks indicate statistical significance after two-sided Mann-Whitney-Wilcoxon tests with Bonferroni correction for multiple comparisons - ns: $0.05 < p$; *: $0.01 < p \leq 0.05$; **: $0.001 < p \leq 0.01$; ***: $0.0001 < p \leq 0.001$; ****: $p \leq 0.0001$.}
    \vspace{-1.5em}
    \label{fig:principal_features}
\end{figure*}

We first plot physiological changes induced by robot encounters in Fig. \ref{fig:over_trial}, represented by 2 principal features: HRV composite multiscale entropy (``HRV\_CMSEn") and the number of EDA phasic SMNA peaks normalized by length of the recording (``nsEDRfreq").
HRV\_CMSEn is calculated from the chest e-tattoo's ECG signals using the methodology in \cite{CMSEn} and nsEDRfreq is calculated from the E4's EDA signals using the methodology in \cite{greco2021acute}. 
HRV\_CMSEn and nsEDRfreq showed high negative correlation ($r = -0.8478$) in their values over experiment time (``trials"), which show successful  estimation of acute stress over time in which 2 devices with different sensing modalities observed similar stress-indicative signal properties across the experiment. This agrees with literature reporting increases in EDA phasic activity (i.e. nsEDRfreq) and decreases in HRV entropy (i.e. HRV\_CMSEn) to be indicative of increased acute stress (see Section \ref{ssec:physiosignals}). \loosepar 
We plotted average feature changes over all participants according to the encounter conditions and conducted pairwise comparisons (Fig. \ref{fig:principal_features}) to investigate if the controlled encounter conditions induced physiological changes compared to baseline across all subjects. 

\subsection{Classification of Human-Robot Encounters} 
\label{sssec:ml}
To confirm the insights from the exploratory analysis, we built a machine learning-based classifier of human-robot encounters using features from EDA and ECG signals. 
To ensure fair evaluation, we report leave-one-subject-out (LOSO) cross-validated results. 
In each iteration of LOSO cross-validation, the model is trained on data from all subjects other than one before prediction is evaluated on data from the unseen subject.
This process is repeated the same number of times as the number of subjects to cover all possible train-test splits. 
LOSO cross-validation therefore ensures that the reported performance accounts for the classifier's ability to predict the type of human-robot encounter experienced by any potential unseen person wearing the same sensors. 
We decided to use random undersampling boosting (RusBoost) \cite{rusboost} for the binary classification and multi-class adaptive boosting (AdaBoost) \cite{adaboostm2} for the multi-class classification. 
For the former, RusBoost was chosen for its known robustness to class imbalances resulting from our study design where subjects spent more time encountering robot(s) rather than not, whereas for the multi-class classification, this was not a concern due to the further categorization of robot encounters into types. 
The maximum number of splits was fixed at 20.\loosepar

\section{Results}
\begin{figure*}[h!]
    \centering
    \includegraphics[width=\textwidth]{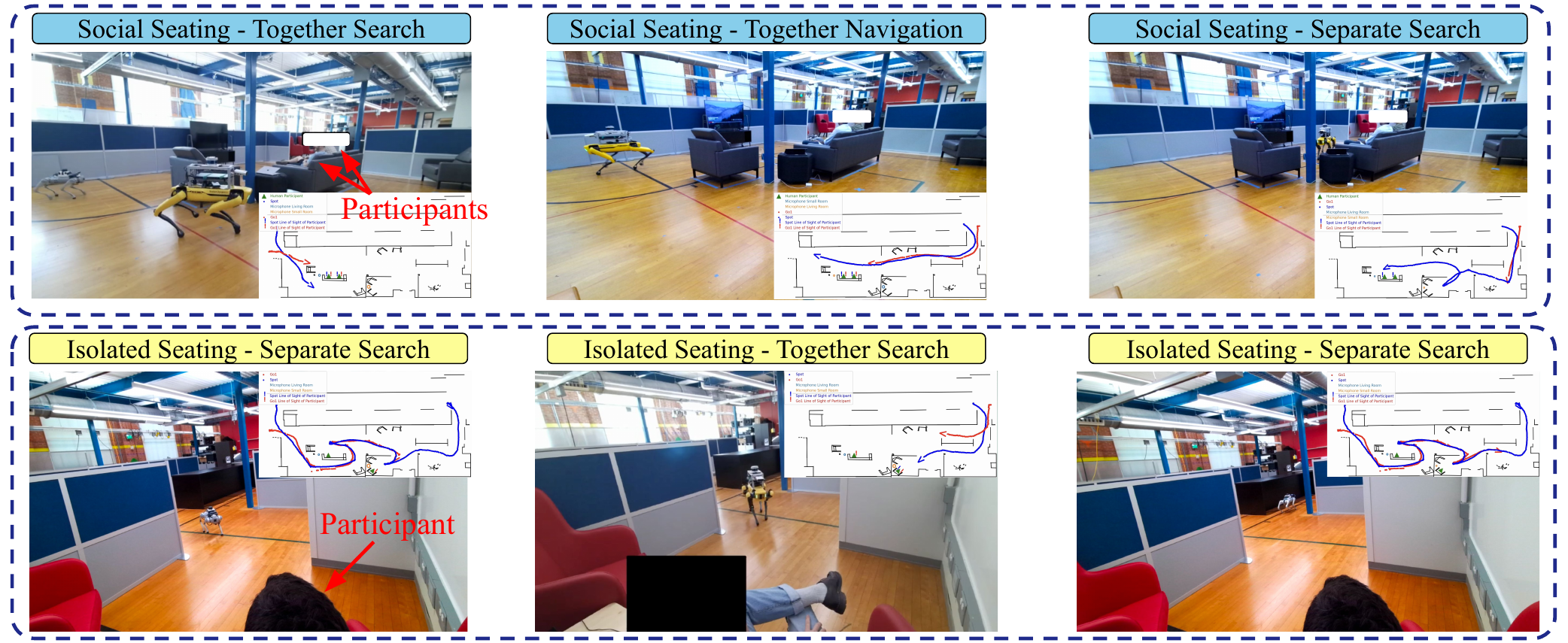}
    \vspace{-2em}
    \caption{Moments captured during encounters with real human participants. Overlays are used to protect the identity of participants. Robot path and position in various behaviors is overlaid on a map of the apartment environment at the moment captured. Robot data is visualized using the XDF files.\loosepar}
    \vspace{-1em}
    \label{fig:mosaic}
\end{figure*}

All of the data taken during experiments including from robots, microphones, and physiological sensors is published for open access at \cite{dataset} and a video of experimental encounters can be found at \url{https://youtu.be/3xu5AoBYi44}.
Quantitative results are also reported in the data repository.
Fig. \ref{fig:mosaic} demonstrates screenshots from several encounters with human participants.
For data from all subjects combined, the Mann-Whitney-Wilcoxon two-sided test was used with Bonferroni correction for multiple comparisons. 
On average, we found that subjects were: more stressed whenever the robots were in the shared environment compared to when they were not; more stressed when the robots were travelling together rather than alone; and more stressed when the robots were navigating compared to searching (Fig. \ref{fig:principal_features}(a-c)). Whether the subjects were located alone or together did not induce significant changes in the stress features examined (Fig. \ref{fig:principal_features}(d)). \loosepar

For subject-specific data (Fig. \ref{fig:over_trial}), to control for inter-subject variability, repeated measures ANOVA was used to investigate the effect of experiment conditions on the trial-level EDA response (nsEDRfreq). As noted on Table \ref{table:rmanova}, we found that the EDA response within subjects was significantly altered across the 10 trials. Specifically, individual subjects had their EDA response change significantly depending on the presence of a robot (Baseline-Robot) and robot behavior.\loosepar

The predictive value of the extracted HRV and EDA features were validated by classifying human-robot encounters via machine learning according to Section \ref{sssec:ml}. The resulting LOSO-validated confusion matrices are presented in Tables \ref{table:cm1} - \ref{table:cm4}. As hinted by the statistical significance of differences in principal features (Fig. \ref{fig:principal_features}) between the encounter conditions, it is within expectation that the binary classifier for identifying robot encounters was the most effective, followed by 3-class classifiers for single- vs. multi-robot encounters and encounters with navigating robots vs. searching robots. Classifying isolated- vs. together-seating encounters was the least effective. \loosepar 

In Fig \ref{fig:principal_features}, the values are adjusted with Bonferroni correction. 
The two-sided Mann-Whitney-Wilcoxon tests with Bonferroni correction that yields pairwise across-subjects significance is different from the p-values reported in Table \ref{table:rmanova}, which instead refers to the within-subject analysis with repeated measures ANOVA. 
Specifically, the p-values in Table \ref{table:rmanova} are the p-values of the F-statistic, which simply shows how likely is it that the null hypothesis of ``no difference among group means" is true. 
Therefore, for this analysis, to the best of our knowledge, adjustment is not necessary.
There is, however, a procedure for instead getting pairwise (e.g. isolated-baseline, social-baseline, social-isolated) adjusted p-values for within-subjects RMANOVA, but it is very complicated and will result in similar information to Fig. \ref{fig:over_trial}, which presents the pairwise comparisons across subjects, which is a more generalizable result than the within-subjects one.

\begin{center}
\begin{threeparttable}
\vspace{-2em}
\caption{Results of the repeated measures ANOVA for the number of phasic EDA responses during each encounter condition}
\begin{tabular}{rlcc}
\hline
&   & $F$ value & $p$-value \\
\hline
&Trials ($df=9$)         & $2.763$ & $0.00588^{**}$ \\ 
&Baseline-Robot ($df=1$)  & $13.477$ & $0.000365^{***}$ \\
&Participant Seating ($df=1$) & $0.463$ & $0.498$ \\
&Robot Behavior ($df=1$)  & $5.986$ & $0.0159^{*}$ \\ 
&Single-Multi Robot ($df=1$)  & $2.215$ & $0.139$ \\
\hline
\end{tabular}
\begin{tablenotes}
\item $^{***}p$-value $\leq 0.001$; $^{**}p$-value $\leq 0.01$; $^{*}p$-value $\leq 0.05$.
\end{tablenotes}
\label{table:rmanova}
\end{threeparttable}
\end{center}



\begin{table}[h]
\begin{center}
\vspace{1em}
\caption{Robot Encounters vs. Baseline}
\begin{tabular}{cc|c|c|c} 
\multicolumn{4}{c}{Predicted} \\ 
\multirow{4}{*}{\rotatebox{90}{Actual}} & & Baseline & Robot \\ \cline{2-4}
    & Baseline & \cellcolor{blue!25}$70.0\%$ & $30.0\%$ \\ \cline{2-4}
    & Robot & $17.5\%$ & \cellcolor{blue!25}$82.5\%$ \\ 
\label{table:cm1}
\vspace{-3em}
\end{tabular}
\end{center}
\end{table}


\begin{table}[h]
\begin{center}
\caption{Single vs. Multi-Robot Encounters}
\begin{tabular}{ cc|c|c|c } 
  \multicolumn{5}{c}{Predicted} \\ 
\multirow{5}{*}{\rotatebox{90}{Actual}} & & Baseline & Single & Multi \\ \cline{2-5}
    & Baseline & \cellcolor{blue!25}$56.7\%$ & $36.7\%$ & $6.7\%$ \\ \cline{2-5}
    & Single & $20.0\%$ & \cellcolor{blue!25}$46.7\%$ & $33.3\%$ \\ \cline{2-5}
    & Multi & $10.0\%$ & $33.3\%$ & \cellcolor{blue!25}$56.7\%$ \\ 
\label{table:cm2}
\vspace{-2.5em}
\end{tabular}
\end{center}
\end{table}


\begin{table}[h!]
\begin{center}
\caption{Encounters with Navigating vs. Searching Robots}
\begin{tabular}{ cc|c|c|c } 
  \multicolumn{5}{c}{Predicted} \\ 
\multirow{5}{*}{\rotatebox{90}{Actual}} & & Baseline & Navigation & Search \\ \cline{2-5}
    & Baseline & \cellcolor{blue!25}$53.3\%$ & $0.0\%$ & $43.3\%$ \\ \cline{2-5}
    & Navigation & $3.3\%$ & \cellcolor{blue!25}$65.0\%$ & $31.7\%$ \\ \cline{2-5}
    & Search & $16.7\%$ & $30.0\%$ & \cellcolor{blue!25}$53.3\%$ \\ 
\label{table:cm3}
\vspace{-2.5em}
\end{tabular}
\end{center}
\end{table}

\begin{table}[h!]
\begin{center}
\caption{Encounters in Isolated vs. Social Seating}
\begin{tabular}{ cc|c|c|c } 
  \multicolumn{5}{c}{Predicted} \\ 
\multirow{5}{*}{\rotatebox{90}{Actual}} & & Baseline & Isolated & Social \\ \cline{2-5}
    & Baseline & \cellcolor{blue!25}$56.7\%$ & $33.3\%$ & $10.0\%$ \\ \cline{2-5}
    & Isolated & $26.8\%$ & \cellcolor{blue!25}$46.4\%$ & $26.8\%$ \\ \cline{2-5}
    & Social & $7.8\%$ & $46.9\%$ & \cellcolor{blue!25}$45.3\%$ \\ 
\label{table:cm4}
\vspace{-3em}
\end{tabular}
\end{center}
\end{table}

\section{Discussion}
Fig. \ref{fig:principal_features} and Tables \ref{table:cm1} - \ref{table:cm4} suggest that for the average human participant, sharing indoor spaces with mobile quadrupedal robots was associated with statistically significant physiological changes indicative of acute stress, which provides a response to $Q1$.
Further, it was possible to distinguish, with higher than chance-level accuracy, specific characteristics (single v. multi robot and navigation v. search) about the human-robot encounters based on the ECG and EDA signals. 
These statistical correlations provide insight to research questions $Q2$.
These observations should be critically examined for potential insights for designers of robots and human-robot interaction.

For instance, we found that the average participant was more stressed by robots on navigation behavior than those on search behavior. 
One could hypothesize that this was due to robots on search inherently spending more time near the participants, giving them an opportunity to become more familiar with the robot. 
Indeed, familiarity, which is influenced by interaction duration and frequency, is one of the major factors behind perceived safety \cite{akalin2023taxonomy}. 
However, to confirm this idea, further analysis of the data may be necessary (e.g. correlate human-robot distance with stress features to decouple navigation/search behavior from distance; interview and questionnaire data collected may contain information about how differently the subjects perceived robots on navigation vs. search). 
While less statistically significant, we also found that the average participant was more stressed encountering multiple robots travelling together rather than alone. 
This could suggest that whenever possible, mobile robots should travel alone, at least in similar physical settings as our experiment (i.e. closed indoors space occupied by humans), to maximize perceived safety. 
We also found that whether the subjects were alone or together was not associated with significant differences in physiological features of stress, responding to $Q3$. 
This could indicate that  the level of stress perecieved by people due to the co-location by mobile robots
may not pose significant difference when they are
deployed in environments where people work alone vs environments where multiple people work alongside each other.\loosepar

Fig. \ref{fig:over_trial} and Table \ref{table:rmanova} suggest that these changes were consistent when examined on a per-participant basis. 
This could potentially allow longer-term tracking of individuals' perceived safety of mobile robots when combined with the appropriate wearable sensing technology and wireless data acquisition hardware, as demonstrated in this study. 
Overall, this study outlined an objective approach to recording human-robot encounters, and investigated factors affecting perceived safety of mobile quadrupeds in indoor environments by analyzing human physiological markers of acute stress. \loosepar

\section{Conclusions}
We presented results found from experimental human-robot encounters in an indoor apartment setting with members of the University of Texas at Austin.
Particularly, we found that participants showed a measurable response to the robots over the baseline, multiple robots compared to a single robot, and under navigation compared with search behavior.
Results were computed using statistical analysis of features found from electrodermal activity and heart rate variability recorded during experiments.
One can hypothesize that the classifier may generalize to other HRI scenarios where participants are seated in a comfortable indoor environment when robots are present. 
Further, this classifier may provide a baseline from which to continue to develop an online classifier for acute mental stress, which may be used to control robot behavior.
The complete open access database can be found at \cite{dataset} and includes a plethora of rich time series data outside of the scope of this manuscript.
For instance, one can manipulate the robot pose information to identify when the robots are in the line-of-sight of participants or the distance to participants. 
This data offer a new context from which to study proxemics.
The dataset also includes audio intensity data from participant seating locations, which can further the investigation of the impacts of sound on physiological response.
Ongoing work looks to combine the presented results with findings from interviews with participants in order to paint a more complete picture of human response to mobile robot encounters.\loosepar

\section{Acknowledgments}

This research was supported in part by NSF Award \#2219236 (GCR: Community Embedded Robotics: Understanding Sociotechnical Interactions with Long-term Autonomous Deployments) and Living and Working with Robots, a core research project of Good Systems, a UT Grand Challenge. Any opinions, findings, and conclusions or recommendations expressed in this material are those of the author and do not necessarily reflect the views of the National Science Foundation.

\printbibliography

@article{gupta2023learned,
  title={Fast LiDAR Informed Visual Search in Unseen Indoor Environments},
  author={Gupta, Ryan and Morgenstein, Kyle and Ortega, Steven and Sentis, Luis},
  journal={arXiv preprint arXiv:2309.14150},
  year={2023}
}

@data{dataset,
author = {Ryan Gupta and Hyonyoung Shin and Emily Norman and Zhiyun Deng and Maria Esteva and Nanshu Lu and Keri K Stephens and Luis Sentis},
publisher = {Texas Data Repository},
title = {{Community Embedded Robotics: A Multimodal Dataset on Perceived Safety during Indoor Mobile Robot Encounters}},
year = {2024},
version = {V1},
doi = {10.18738/T8/FT9VYS},
url = {https://doi.org/10.18738/T8/FT9VYS}
}

@article{wang2023effects,
  title={The effects of natural sounds and proxemic distances on the perception of a noisy domestic flying robot},
  author={Wang, Ziming and Hu, Ziyi and Rohles, Bj{\"o}rn and Ljungblad, Sara and Koenig, Vincent and Fjeld, Morten},
  journal={ACM Transactions on Human-Robot Interaction},
  volume={12},
  number={4},
  pages={1--32},
  year={2023},
  publisher={ACM New York, NY}
}

@article{fortunati2021rise,
  title={The rise of the roboid},
  author={Fortunati, Leopoldina and Sorrentino, Alessandra and Fiorini, Laura and Cavallo, Filippo},
  journal={International Journal of Social Robotics},
  volume={13},
  pages={1457--1471},
  year={2021},
  publisher={Springer}
}

@article{caruelle2019use,
  title={The use of electrodermal activity (EDA) measurement to understand consumer emotions--A literature review and a call for action},
  author={Caruelle, Delphine and Gustafsson, Anders and Shams, Poja and Lervik-Olsen, Line},
  journal={Journal of Business Research},
  volume={104},
  pages={146--160},
  year={2019},
  publisher={Elsevier}
}

@article{critchley2002electrodermal,
  title={Electrodermal responses: what happens in the brain},
  author={Critchley, Hugo D},
  journal={The Neuroscientist},
  volume={8},
  number={2},
  pages={132--142},
  year={2002},
  publisher={SAGE Publications Sage CA: Los Angeles, CA}
}

@article{zangroniz2017electrodermal,
  title={Electrodermal activity sensor for classification of calm/distress condition},
  author={Zangr{\'o}niz, Roberto and Mart{\'\i}nez-Rodrigo, Arturo and Pastor, Jos{\'e} Manuel and L{\'o}pez, Mar{\'\i}a T and Fern{\'a}ndez-Caballero, Antonio},
  journal={Sensors},
  volume={17},
  number={10},
  pages={2324},
  year={2017},
  publisher={MDPI}
}

@article{terry2017thematic,
  title={Thematic analysis},
  author={Terry, Gareth and Hayfield, Nikki and Clarke, Victoria and Braun, Virginia and others},
  journal={The SAGE handbook of qualitative research in psychology},
  volume={2},
  number={17-37},
  pages={25},
  year={2017},
  publisher={SAGE Publishers}
}

@article{joffe2011thematic,
  title={Thematic analysis},
  author={Joffe, Helene},
  journal={Qualitative research methods in mental health and psychotherapy: A guide for students and practitioners},
  pages={209--223},
  year={2011},
  publisher={Wiley Online Library}
}

@article{doring2019love,
  title={Love and sex with robots: a content analysis of media representations},
  author={D{\"o}ring, Nicola and Poeschl, Sandra},
  journal={International Journal of Social Robotics},
  volume={11},
  number={4},
  pages={665--677},
  year={2019},
  publisher={Springer}
}

@article{mavrogiannis2022social,
  title={Social momentum: Design and evaluation of a framework for socially competent robot navigation},
  author={Mavrogiannis, Christoforos and Alves-Oliveira, Patr{\'\i}cia and Thomason, Wil and Knepper, Ross A},
  journal={ACM Transactions on Human-Robot Interaction (THRI)},
  volume={11},
  number={2},
  pages={1--37},
  year={2022},
  publisher={ACM New York, NY}
}

@article{aliasghari2023we,
  title={How do we perceive our trainee robots? exploring the impact of robot errors and appearance when performing domestic physical tasks on teachers’ trust and evaluations},
  author={Aliasghari, Pourya and Ghafurian, Moojan and Nehaniv, Chrystopher L and Dautenhahn, Kerstin},
  journal={ACM Transactions on Human-Robot Interaction},
  volume={12},
  number={3},
  pages={1--41},
  year={2023},
  publisher={ACM New York, NY}
}

@article{choi2005peer,
  title={Peer reviewed: a catalog of biases in questionnaires},
  author={Choi, Bernard CK and Pak, Anita WP},
  journal={Preventing chronic disease},
  volume={2},
  number={1},
  year={2005},
  publisher={Centers for Disease Control and Prevention}
}

@article{akalin2022you,
  title={Do you feel safe with your robot? Factors influencing perceived safety in human-robot interaction based on subjective and objective measures},
  author={Akalin, Neziha and Kristoffersson, Annica and Loutfi, Amy},
  journal={International journal of human-computer studies},
  volume={158},
  pages={102744},
  year={2022},
  publisher={Elsevier}
}

@inproceedings{chen2011touched,
  title={Touched by a robot: An investigation of subjective responses to robot-initiated touch},
  author={Chen, Tiffany L and King, Chih-Hung and Thomaz, Andrea L and Kemp, Charles C},
  booktitle={Proceedings of the 6th international conference on Human-robot interaction},
  pages={457--464},
  year={2011}
}

@article{dehais2011physiological,
  title={Physiological and subjective evaluation of a human--robot object hand-over task},
  author={Dehais, Fr{\'e}d{\'e}ric and Sisbot, Emrah Akin and Alami, Rachid and Causse, Micka{\"e}l},
  journal={Applied ergonomics},
  volume={42},
  number={6},
  pages={785--791},
  year={2011},
  publisher={Elsevier}
}

@inproceedings{stark2018personal,
  title={Personal space intrusion in human-robot collaboration},
  author={Stark, Jessi and Mota, Roberta RC and Sharlin, Ehud},
  booktitle={Companion of the 2018 ACM/IEEE international conference on human-robot interaction},
  pages={245--246},
  year={2018}
}

@article{willemse2017affective,
  title={Affective and behavioral responses to robot-initiated social touch: toward understanding the opportunities and limitations of physical contact in human--robot interaction},
  author={Willemse, Christian JAM and Toet, Alexander and Van Erp, Jan BF},
  journal={Frontiers in ICT},
  volume={4},
  pages={12},
  year={2017},
  publisher={Frontiers Media SA}
}

@article{rubagotti2022perceived,
  title={Perceived safety in physical human--robot interaction—A survey},
  author={Rubagotti, Matteo and Tusseyeva, Inara and Baltabayeva, Sara and Summers, Danna and Sandygulova, Anara},
  journal={Robotics and Autonomous Systems},
  volume={151},
  pages={104047},
  year={2022},
  publisher={Elsevier}
}

@article{akalin2023taxonomy,
  title={A Taxonomy of Factors Influencing Perceived Safety in Human--Robot Interaction},
  author={Akalin, Neziha and Kiselev, Andrey and Kristoffersson, Annica and Loutfi, Amy},
  journal={International Journal of Social Robotics},
  volume={15},
  number={12},
  pages={1993--2004},
  year={2023},
  publisher={Springer}
}

@article{turner2019use,
  title={Use of a non-human robot audience to induce stress reactivity in human participants},
  author={Turner-Cobb, Julie M and Asif, Mashal and Turner, James E and Bevan, Chris and Fraser, Danae Stanton},
  journal={Computers in Human Behavior},
  volume={99},
  pages={76--85},
  year={2019},
  publisher={Elsevier}
}

@inproceedings{van2023increasing,
  title={Increasing perceived safety in motion planning for human-drone interaction},
  author={Van Waveren, Sanne and Rudling, Rasmus and Leite, Iolanda and Jensfelt, Patric and Pek, Christian},
  booktitle={Proceedings of the 2023 ACM/IEEE International Conference on Human-Robot Interaction},
  pages={446--455},
  year={2023}
}

@inproceedings{taylor2022observer,
  title={Observer-aware legibility for social navigation},
  author={Taylor, Ada V and Mamantov, Ellie and Admoni, Henny},
  booktitle={2022 31st IEEE International Conference on Robot and Human Interactive Communication (RO-MAN)},
  pages={1115--1122},
  year={2022},
  organization={IEEE}
}

@article{bhattacharya2023chest,
  title={A Chest-Conformable, Wireless Electro-Mechanical E-Tattoo for Measuring Multiple Cardiac Time Intervals},
  author={Bhattacharya, Sarnab and Nikbakht, Mohammad and Alden, Alec and Tan, Philip and Wang, Jieting and Alhalimi, Taha A and Kim, Sangjun and Wang, Pulin and Tanaka, Hirofumi and Tandon, Animesh and others},
  journal={Advanced Electronic Materials},
  pages={2201284},
  year={2023},
  publisher={Wiley Online Library}
}

@inproceedings{sikand2021robofleet,
  title={Robofleet: Open source communication and management for fleets of autonomous robots},
  author={Sikand, Kavan Singh and Zartman, Logan and Rabiee, Sadegh and Biswas, Joydeep},
  booktitle={2021 IEEE/RSJ International Conference on Intelligent Robots and Systems (IROS)},
  pages={406--412},
  year={2021},
  organization={IEEE}
}

@misc{lsl, url={https://github.com/sccn/labstreaminglayer}}

@misc{labrecorder, url={https://github.com/labstreaminglayer/App-LabRecorder}}

@article{greco2015cvxeda,
  title={cvxEDA: A convex optimization approach to electrodermal activity processing},
  author={Greco, Alberto and Valenza, Gaetano and Lanata, Antonio and Scilingo, Enzo Pasquale and Citi, Luca},
  journal={IEEE transactions on biomedical engineering},
  volume={63},
  number={4},
  pages={797--804},
  year={2015},
  publisher={IEEE}
}

@article{greco2021acute,
  title={Acute stress state classification based on electrodermal activity modeling},
  author={Greco, Alberto and Valenza, Gaetano and Lazaro, Jesus and Garzon-Rey, Jorge Mario and Aguilo, Jordi and De-la-Camara, Concepcion and Bailon, Raquel and Scilingo, Enzo Pasquale},
  journal={IEEE Transactions on Affective Computing},
  year={2021},
  publisher={IEEE}
}

@article{grocholsky2006cooperative,
  title={Cooperative air and ground surveillance},
  author={Grocholsky, Ben and Keller, James and Kumar, Vijay and Pappas, George},
  journal={IEEE Robotics \& Automation Magazine},
  volume={13},
  number={3},
  pages={16--25},
  year={2006},
  publisher={IEEE}
}

@inproceedings{charrow2015information,
  title={Information-theoretic mapping using cauchy-schwarz quadratic mutual information},
  author={Charrow, Benjamin and Liu, Sikang and Kumar, Vijay and Michael, Nathan},
  booktitle={2015 IEEE International Conference on Robotics and Automation (ICRA)},
  pages={4791--4798},
  year={2015},
  organization={IEEE}
}

@article{lee2021assistive,
  title={Assistive delivery robot application for real-world postal services},
  author={Lee, Daegyu and Kang, Gyuree and Kim, Boseong and Shim, D Hyunchul},
  journal={IEEE Access},
  volume={9},
  pages={141981--141998},
  year={2021},
  publisher={IEEE}
}

@misc{e4, url={https://www.empatica.com/research/e4/}}

@misc{ros.org, url={http://wiki.ros.org/costmap_2d}, journal={ros.org}}

@article{bartneck2009measurement,
  title={Measurement instruments for the anthropomorphism, animacy, likeability, perceived intelligence, and perceived safety of robots},
  author={Bartneck, Christoph and Kuli{\'c}, Dana and Croft, Elizabeth and Zoghbi, Susana},
  journal={International journal of social robotics},
  volume={1},
  pages={71--81},
  year={2009},
  publisher={Springer}
}

@inproceedings{staffa2022enhancing,
  title={Enhancing Affective Robotics via Human Internal State Monitoring},
  author={Staffa, Mariacarla and Rossi, Silvia},
  booktitle={2022 31st IEEE International Conference on Robot and Human Interactive Communication (RO-MAN)},
  pages={884--890},
  year={2022},
  organization={IEEE}
}

@article{rani2004anxiety,
  title={Anxiety detecting robotic system--towards implicit human-robot collaboration},
  author={Rani, Pramila and Sarkar, Nilanjan and Smith, Craig A and Kirby, Leslie D},
  journal={Robotica},
  volume={22},
  number={1},
  pages={85--95},
  year={2004},
  publisher={Cambridge University Press}
}

@article{kim2018stress,
  title={Stress and heart rate variability: a meta-analysis and review of the literature},
  author={Kim, Hye-Geum and Cheon, Eun-Jin and Bai, Dai-Seg and Lee, Young Hwan and Koo, Bon-Hoon},
  journal={Psychiatry investigation},
  volume={15},
  number={3},
  pages={235},
  year={2018},
  publisher={Korean Neuropsychiatric Association}
}

@inproceedings{hauser2023s,
  title={" What’s That Robot Doing Here?": Perceptions Of Incidental Encounters With Autonomous Quadruped Robots},
  author={Hauser, Elliott and Chan, Yao-Cheng and Chonkar, Parth and Hemkumar, Geethika and Wang, Huihai and Dua, Daksh and Gupta, Shikhar and Enriquez, Efren Mendoza and Kao, Tiffany and Hart, Justin and others},
  booktitle={Proceedings of the First International Symposium on Trustworthy Autonomous Systems},
  pages={1--15},
  year={2023}
}

@article{cohen1994perceived,
  title={Perceived stress scale},
  author={Cohen, Sheldon and Kamarck, Tom and Mermelstein, Robin and others},
  journal={Measuring stress: A guide for health and social scientists},
  volume={10},
  number={2},
  pages={1--2},
  year={1994}
}

@article{makowski2021neurokit2,
  title={NeuroKit2: A Python toolbox for neurophysiological signal processing},
  author={Makowski, Dominique and Pham, Tam and Lau, Zen J and Brammer, Jan C and Lespinasse, Fran{\c{c}}ois and Pham, Hung and Sch{\"o}lzel, Christopher and Chen, SH Annabel},
  journal={Behavior research methods},
  pages={1--8},
  year={2021},
  publisher={Springer}
}

@article{thehrvreview,
    author = {Immanuel, Sarah and Teferra, Meseret N. and Baumert, Mathias and Bidargaddi, Niranjan},
    title = "{Heart Rate Variability for Evaluating Psychological Stress Changes in Healthy Adults: A Scoping Review}",
    journal = {Neuropsychobiology},
    volume = {82},
    number = {4},
    pages = {187-202},
    year = {2023},
    month = {06},
    issn = {0302-282X}
}

@article{USTHRV1, title={Ultra short term analysis of heart rate variability for monitoring mental stress in Mobile settings}, journal={2007 29th Annual International Conference of the IEEE Engineering in Medicine and Biology Society}, author={Salahuddin, Lizawati and Cho, Jaegeol and Jeong, Myeong Gi and Kim, Desok}, year={2007}, month={Aug}}

@article{USTHRV2, title={Mental stress assessment using ultra short term HRV analysis based on non-linear method}, volume={12}, number={7}, journal={Biosensors}, author={Lee, Seungjae and Hwang, Ho Bin and Park, Seongryul and Kim, Sanghag and Ha, Jung Hee and Jang, Yoojin and Hwang, Sejin and Park, Hoon-Ki and Lee, Jongshill and Kim, In Young}, year={2022}, month={Jun}, pages={465}}

@article{E4popularity, title={Wearables measuring electrodermal activity to assess perceived stress in care: a scoping review}, journal={Acta Neuropsychiatrica}, author={Klimek, Agata and Mannheim, Ittay and Schouten, Gerard and Wouters, Eveline J. M. and Peeters, Manon W. H.}, year={2023}, pages={1–11}}

@article{brugnera_entropy,
title = {Heart rate variability during acute psychosocial stress: A randomized cross-over trial of verbal and non-verbal laboratory stressors},
journal = {International Journal of Psychophysiology},
volume = {127},
pages = {17-25},
year = {2018},
issn = {0167-8760},
author = {Agostino Brugnera and Cristina Zarbo and Mika P. Tarvainen and Paolo Marchettini and Roberta Adorni and Angelo Compare}
}

@article{dmitri_entropy, title={State anxiety and nonlinear dynamics of Heart Rate Variability in students}, volume={11}, number={1}, journal={PLOS ONE}, author={Dimitriev, Dimitriy A. and Saperova, Elena V. and Dimitriev, Aleksey D.}, year={2016}, month={Jan}}

@INPROCEEDINGS{rusboost,
  author={Seiffert, Chris and Khoshgoftaar, Taghi M. and Van Hulse, Jason and Napolitano, Amri},
  booktitle={2008 19th International Conference on Pattern Recognition}, 
  title={RUSBoost: Improving classification performance when training data is skewed}, 
  year={2008},
  volume={},
  number={},
  pages={1-4},
}

@article{adaboostm2,
  title={Multi-class adaboost},
  author={Hastie, Trevor and Rosset, Saharon and Zhu, Ji and Zou, Hui},
  journal={Statistics and its Interface},
  volume={2},
  number={3},
  pages={349--360},
  year={2009},
  publisher={International Press of Boston}
}

@article{exam_entropy, title={Nonlinear heart rate variability features for real-life stress detection. case study: Students under stress due to university examination}, volume={10}, number={1}, journal={BioMedical Engineering OnLine}, author={Melillo, Paolo and Bracale, Marcello and Pecchia, Leandro}, year={2011}, month={Nov}, pages={96}}

@Article{CMSEn,
AUTHOR = {Wu, Shuen-De and Wu, Chiu-Wen and Lin, Shiou-Gwo and Wang, Chun-Chieh and Lee, Kung-Yen},
TITLE = {Time Series Analysis Using Composite Multiscale Entropy},
JOURNAL = {Entropy},
VOLUME = {15},
YEAR = {2013},
NUMBER = {3},
PAGES = {1069--1084},
URL = {https://www.mdpi.com/1099-4300/15/3/1069},
ISSN = {1099-4300}
}

@article{autocor,
author = {van Halem, Sjoerd and van Roekel, Eeske and Kroencke, Lara and Kuper, Niclas and Denissen, Jaap},
title = {Moments That Matter? On the Complexity of Using Triggers Based on Skin Conductance to Sample Arousing Events Within an Experience Sampling Framework},
journal = {European Journal of Personality},
volume = {34},
number = {5},
pages = {794-807},
year = {2020}
}

\end{document}